\documentclass{article}

\usepackage{arxiv}

\usepackage[utf8]{inputenc} 
\usepackage[T1]{fontenc}    
\usepackage{url}            
\usepackage{booktabs}       
\usepackage{amsfonts}       
\usepackage{amsmath}
\usepackage{nicefrac}       
\usepackage{microtype}      
\usepackage{graphicx}
\usepackage{xcolor}
\usepackage{tabularx}
\usepackage{makecell}
\usepackage{pdflscape}      
\usepackage{fancyvrb}       
\usepackage[font=small,labelfont=bf,labelsep=period]{caption}
\usepackage[numbers,square,sort&compress]{natbib}
\usepackage{hyperref}       
\usepackage{doi}

\hypersetup{colorlinks=true, linkcolor=blue!50!black, citecolor=blue!50!black, urlcolor=blue!50!black}

\newcolumntype{L}{>{\raggedright\arraybackslash}X}

\setcellgapes{1pt}
\newcommand{\tabnote}[1]{\par\smallskip\begin{minipage}{\linewidth}\footnotesize\raggedright #1\end{minipage}}

\title{Jev in Medicine: A Benchmark Evaluation}

\author{%
  \href{https://orcid.org/0000-0002-1591-0467}{\includegraphics[scale=0.06]{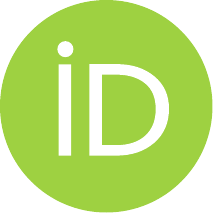}\hspace{1mm}Alfredo Madrid-García}\thanks{Corresponding author.} \\
  Independent researcher \\
  Madrid, Spain \\
  \texttt{fredymad@msn.com} \\
  \And
  \href{https://orcid.org/0000-0001-5070-4178}{\includegraphics[scale=0.06]{orcid.pdf}\hspace{1mm}Beatriz Merino-Barbancho} \\
  Independent researcher \\
  Madrid, Spain \\
}

\renewcommand{\headeright}{Preliminary results}
\renewcommand{\undertitle}{A Preprint}
\renewcommand{\shorttitle}{Jev in Medicine: A Benchmark Evaluation}

\hypersetup{
  pdftitle={Jev in Medicine: A Benchmark Evaluation},
  pdfsubject={cs.CL, cs.AI},
  pdfauthor={Alfredo Madrid-García, Beatriz Merino-Barbancho},
  pdfkeywords={Artificial intelligence, Large language models, System One model, Calibration, Clinical decision-making, Benchmarking, Jev},
}

\begin{document}
\maketitle

\begin{abstract}
\textbf{Background:} Jev is a non-generative System One model that assigns probabilities to predefined answer options and cannot answer outside them. Its calibration and its performance on complex diagnostic cases remain insufficiently characterised.

\textbf{Methods:} We evaluated Jev 1.13 on four medical benchmarks: MetaMedQA (1,373 examination questions), PubMedQA (500 research questions), DiagnosisArena-MCQ (915 cases) and the NEJM Case Challenges (34 cases). GPT-6 Sol, with and without reasoning, was the reference. Outcomes were top-1 accuracy (primary), calibration, selective prediction and recognition of unanswerable questions.

\textbf{Results:} Jev's accuracy was similar to that of GPT-6 Sol with medium reasoning on PubMedQA (78.4\% vs 78.2\%), lower on MetaMedQA (74.8\% vs 82.7\%) and much lower on DiagnosisArena-MCQ (59.8\% vs 82.4\%) and the NEJM cases (61.8\% vs 82.4\%, adjusted P = 0.078). On MetaMedQA, Jev's probabilities were the best calibrated (expected calibration error 0.063 vs 0.146), and its answers with a probability $\geq$0.9 (52.9\% of questions) were 93.4\% accurate, but GPT-6 Sol was as accurate when it accepted a similar proportion of questions. On DiagnosisArena-MCQ, Jev's probabilities discriminated poorly (AUROC 0.645 vs 0.768). Of the 162 questions whose correct answer was ``I don't know or cannot answer'', Jev chose that option for 10.5\% (GPT-6 Sol, 8.6\%). Median latency was 0.27--0.31 s; all 2,823 items cost US\$0.08.

\textbf{Conclusions:} Jev was fast and inexpensive, and its accuracy was similar to that of a frontier LLM on research abstracts but lower on examination questions and much lower on complex diagnostic cases. Its better calibration on examination questions gave no selective-prediction advantage, and it seldom chose ``I don't know'' when that was the correct answer. Task-specific validation is required before clinical use.

\textbf{NOTE:} These findings are preliminary. Independent replication, clinical adjudication of reference answers, and assessment of run-to-run stability remain pending. The results may change after further verification and should not be used to guide clinical care.
\end{abstract}


\section{Introduction}\label{sec:intro}

\begin{figure}[tb]
\centering
\includegraphics[width=\linewidth]{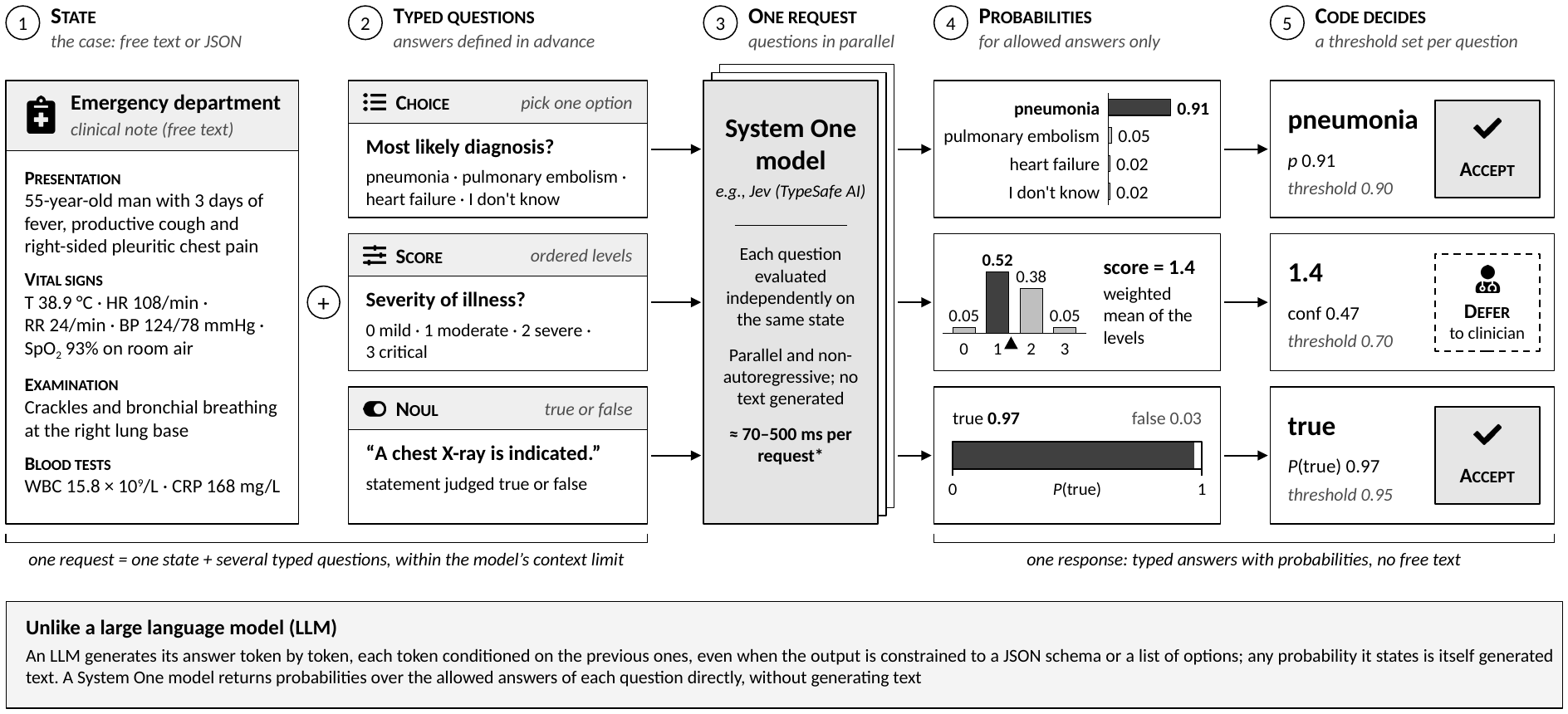}
\caption{How a System One model works.}\label{fig:1}
\end{figure}

Jev, released in early access by TypeSafe AI on 15 September 2026, is the company's first ``System One'' model~\citep{Almeida2026,TypeSafeAI2026docs}. Unlike large language models (LLMs), which build each answer token by token, Jev does not generate text. Instead, it reads the whole input at once and returns all its outputs in a single, non-autoregressive pass. It receives a ``state'', i.e., free text or structured data describing a case, together with one or more typed questions (Figure~\ref{fig:1}). Its answers are restricted to options the user defines in advance through three question types: Choice (one option among up to 255), Score (a position on an ordered rubric) and Noul (the probability that a statement is true). Answers are returned as probabilities, with a separate confidence value for Choice and Score, and the model is post-trained with Reinforcement Learning for Calibrated Decisions to make these probabilities match observed frequencies rather than to produce text that humans prefer~\citep{Almeida2026,TypeSafeAI2026docs}. Its reported end-to-end response times are 70--500 ms, and it costs US\$0.042 per million input tokens, with no charge for outputs. Because its answers cannot fall outside the predefined options, it is also said to be unable to hallucinate.

In medicine, LLMs are being explored for many clinical tasks and reach passing thresholds in licensing and specialty examinations~\citep{Thirunavukarasu2023,MadridGarcia2023,Singhal2023}. They can, however, fabricate content~\citep{Asgari2025}, make errors when free text must be converted into structured decisions~\citep{Hager2024,Soroush2024} and communicate uncertainty poorly~\citep{Savage2025,Griot2025}, and they are costly at health-system scale~\citep{Klang2024}. Many applications of artificial intelligence (AI) in medicine require a choice among predefined alternatives, such as selecting the most likely diagnosis, assigning a triage level or a diagnostic code, or deciding whether a patient meets the eligibility criteria of a clinical trial~\citep{Jin2024} or an article meets the inclusion criteria of a systematic review~\citep{Guo2024}. For these tasks, Jev could have four advantages. First, it cannot return malformed answers or non-existent options, although it must select one of the options provided even when none is correct and can abstain explicitly only if an option such as ``I don't know'' is offered. Second, if its probabilities are calibrated, confident answers could be acted on and uncertain ones referred to a clinician; in other diagnostic AI systems, this strategy has improved accuracy and reduced clinician workload~\citep{Dvijotham2023}. Third, complex judgements can be broken down into atomic questions and combined by explicit rules, as clinical scores do, so that each step can be audited. Fourth, its speed and cost could make it feasible to apply AI-supported decisions to every record, message or article in a health system, or to verify the outputs of generative models or agents.

Evidence on Jev's performance in medicine is scarce. A preprint used Jev to judge the factual agreement between generated and reference radiology reports; its scores correlated with expert error counts (Kendall's $\tau$, 0.40--0.57), but the authors recommended recalibrating its probabilities for each task~\citep{Huang2026}. An exploratory benchmark evaluated Jev on tabular classification tasks, including the Breast Cancer Wisconsin (Diagnostic) dataset~\citep{statsguysam2026}, and an open-source project has adapted its question format to extract clinical variables from free-text notes~\citep{Ma2026}. Finally, an exploratory evaluation reported competitive medical question-answering performance~\citep{Pierri2026}, but Jev's calibration and performance on complex diagnostic cases remain insufficiently characterised. Clinical cases require multi-step reasoning over numerical data and irrelevant findings, all three of which the developer lists as weaknesses of the model~\citep{TypeSafeAI2026jagged}. Benchmark performance alone cannot establish clinical utility or safety, but it is a necessary step before any clinical evaluation.

The objective of this study was therefore to evaluate the accuracy and calibration of Jev on four complementary medical benchmarks, using a single Choice question per item, and to compare it with GPT-6 Sol, a frontier reasoning LLM, run with and without reasoning. The benchmarks were MetaMedQA, which includes unanswerable questions~\citep{Griot2025}; PubMedQA, which requires answering research questions from biomedical abstracts~\citep{Jin2019}; DiagnosisArena-MCQ, a set of challenging published cases~\citep{Zhu2026}; and the New England Journal of Medicine (NEJM) Case Challenges, which also allowed comparison with the journal's readers~\citep{Eriksen2024}. We also explored the recognition of unanswerable questions, selective prediction, response time and cost.

\section{Material and methods}\label{sec:methods}

\subsection{Study design}\label{sec:design}

We conducted a benchmarking study of Jev on four existing medical datasets, with GPT-6 Sol (OpenAI) evaluated on the same items as a reference. The analysis was organised around three questions: how accurate Jev is across tasks of increasing clinical complexity (Q1); whether its probabilities are informative enough to decide which answers to accept and which to defer to a clinician (Q2); and whether it recognises questions that cannot be answered (Q3).

\subsection{Benchmarks}\label{sec:benchmarks}

The four benchmarks span a gradient from the recall of medical knowledge to diagnostic reasoning in complex published cases (Table~\ref{tab:1}). MetaMedQA extends United States Medical Licensing Examination (USMLE)-style questions from MedQA with questions modified so that no substantive option is correct, malformed questions and questions about a fictional organ~\citep{Jin2021}. For PubMedQA, we used the expert-annotated test split and removed the conclusions of each abstract, which reveal the answer. For DiagnosisArena-MCQ, we used the publicly released test cases. Finally, we used a personal subscription to retrieve the 35 NEJM Case Challenges published between May 2020 and September 2026, each of which ends with a reader poll among six candidate diagnoses. Of these, 34 had a published final diagnosis and were scored, and 33 also had a closed poll. Because Jev accepts text only, figures were replaced by their captions and laboratory tables were transcribed as text. All items were otherwise used as published, and reference answers were not re-adjudicated; for the NEJM cases, the correct option was matched by hand to the final diagnosis published in the Case Record (Supplementary~\ref{sec:S1}).

\begin{table}[tbp]
\caption{Benchmarks evaluated.}\label{tab:1}
\centering\small
\begin{tabularx}{\linewidth}{@{}>{\raggedright\arraybackslash}p{2.35cm}>{\raggedright\arraybackslash}p{1.45cm}LLL@{}}
\toprule
Benchmark & Items analysed & Answer options & Task & Distinctive feature\\
\midrule
MetaMedQA~\citep{Griot2025} & 1,373 & 6: A--D, ``None of the above'', ``I don't know or cannot answer'' & USMLE-style examination questions & 277 questions without a correct substantive option (115 ``None of the above'', 162 ``I don't know''), including 100 on a fictional organ\\
\addlinespace[4pt]
PubMedQA~\citep{Jin2019} & 500 & 3: yes, no, maybe & Answer a research question from a PubMed abstract (conclusions removed) & Expert-annotated official test split; single-annotator accuracy 78.0\%\\
\addlinespace[4pt]
DiagnosisArena-MCQ~\citep{Zhu2026} & 915 & 4: A--D & Most likely diagnosis in cases from case reports in ten high-impact journals & Cases filtered to remove those that LLMs solved easily; distractors derived from errors of reasoning models\\
\addlinespace[4pt]
NEJM Case Challenges~\citep{Eriksen2024} & 34 (33 with a closed poll) & 6: A--F & Most likely diagnosis in Case Records of the Massachusetts General Hospital & 273,362 reader votes; laboratory tables transcribed as text, figures replaced by captions\\
\bottomrule
\end{tabularx}
\tabnote{All items were evaluated with Jev 1.13 and with GPT-6 Sol with medium reasoning and without reasoning. Of the 35 NEJM Case Challenges retrieved, one had no published final diagnosis and was not scored. Sources, licences and preparation are described in Supplementary~\ref{sec:S1}. LLM, large language model; MCQ, multiple-choice question; USMLE, United States Medical Licensing Examination.}
\end{table}

\subsection{Models}\label{sec:models}

Because registration on the developer's own platform was closed at the time of the study, Jev 1.13 was accessed through OpenRouter, which forwards every request for this model to TypeSafe~\citep{OpenRouter2026}. Its knowledge cut-off has not been reported. For each request, Jev returns the selected option and a native probability for every option.

GPT-6 Sol (knowledge cut-off 20 April 2026) was accessed through the OpenAI API at two reasoning-effort levels~\citep{OpenAI2026}. Medium, the default, was the primary comparator. None, in which no reasoning tokens are generated, was included to estimate the contribution of explicit reasoning, although its answer is still generated token by token. The output was constrained by a JSON schema to one of the available options plus a probability for each option. These verbalised probabilities are elicited by the prompt rather than produced natively, so differences in probability quality between the two models were interpreted as differences between the systems as deployed. Token log-probabilities, requested as an exploratory alternative in the none condition, were incomplete and are reported only in the Supplementary Material (\ref{sec:S3} and~\ref{sec:S4}).

\subsection{Evaluation procedure}\label{sec:procedure}

For Jev, the case content formed the ``state'' and the answer options formed the criteria of a single Choice question; for GPT-6 Sol, the same instruction, content and options were placed in one user message, without a system prompt or examples. Prompts were not optimised (Supplementary~\ref{sec:S2}). Jev was used out of the box, with one question per item, rather than with the task decomposition and input preparation its developer recommends. Each model and condition was run once per benchmark, between 26 and 27 September 2026.

\subsection{Outcomes and statistical analysis}\label{sec:stats}

The primary outcome (Q1) was top-1 accuracy in each benchmark, with Wilson 95\% confidence intervals (CIs); items without a valid response, including refusals, would have been counted as incorrect. Accuracy was scored from the answer returned by the API, without replacing it with the highest-probability option. The main contrast compared Jev with GPT-6 Sol with medium reasoning on the same items, using the paired difference in accuracy with a 95\% CI from 2,000 bootstrap resamples and the exact McNemar test; the four P values were adjusted with the Holm method. Comparisons with the none condition were secondary.

For Q2, calibration, discrimination and selective prediction were based on the probability of the selected option, and the Brier score on the complete probability vector; Jev's separate confidence value~\citep{TypeSafeAI2026docs}, which the developer derives from the option probabilities, was treated as a ranking score and compared with the selected-option probability in a supplementary analysis. Calibration was summarised with reliability diagrams and the expected calibration error (ECE; ten equal-width bins), discrimination between correct and incorrect answers with the area under the receiver operating characteristic curve (AUROC), and overall probability quality with the multiclass Brier score (range 0--2). In selective prediction, answers at or above a probability threshold are accepted and the rest are deferred. We characterised it by coverage (the proportion of items accepted) and by the accuracy of accepted answers at thresholds of 0.5, 0.8 and 0.9, which were not tuned on the test data. Because a fixed threshold yields different coverage for different models, they were also compared across all thresholds with accuracy--coverage curves and with the area under the risk--coverage curve (AURC; lower is better)~\citep{Ding2020}. The 95\% CIs for the paired differences in ECE, AUROC and AURC were obtained from 2,000 bootstrap resamples. Because calibration estimates are unstable in small samples~\citep{VanCalster2019}, ECE, AUROC, selective prediction and AURC were not computed for the 34 NEJM cases, for which only the mean selected-option probability and the Brier score are reported.

For Q3, we report for MetaMedQA the accuracy on questions with a correct substantive option, the missing-answer recall (the proportion of questions whose correct answer is ``None of the above'' for which that option was selected) and the unknown recall (the same proportion for ``I don't know or cannot answer'').

Two contextual analyses were descriptive. Each model was compared with NEJM readers as the mean paired difference, across the 33 polled cases, between the model's correctness and the proportion of reader votes for the correct option (95\% CI from 10,000 case-bootstrap resamples). Client-side latency and cost were reported under each model's own access route (Jev, one request at a time; GPT-6 Sol, four concurrent requests) and were not compared head to head. Tests were two-sided with $\alpha$ = 0.05; full definitions are given in Supplementary~\ref{sec:S3}.

\section{Results}\label{sec:results}

\subsection{Completeness of the runs}\label{sec:completeness}

All 12 runs (four benchmarks, three model conditions) were complete. Each of the 8,469 requests returned a valid answer at the first attempt, with no refusals, retries or unusable probability vectors, so every item of MetaMedQA, PubMedQA and DiagnosisArena-MCQ and all 34 scored NEJM cases entered the analyses. In three responses, two from Jev and one from GPT-6 Sol without reasoning, the returned answer was not the option with the highest probability.

\subsection{Accuracy across task types (Q1)}\label{sec:acc}

\begin{figure}[tb]
\centering
\includegraphics[width=\linewidth]{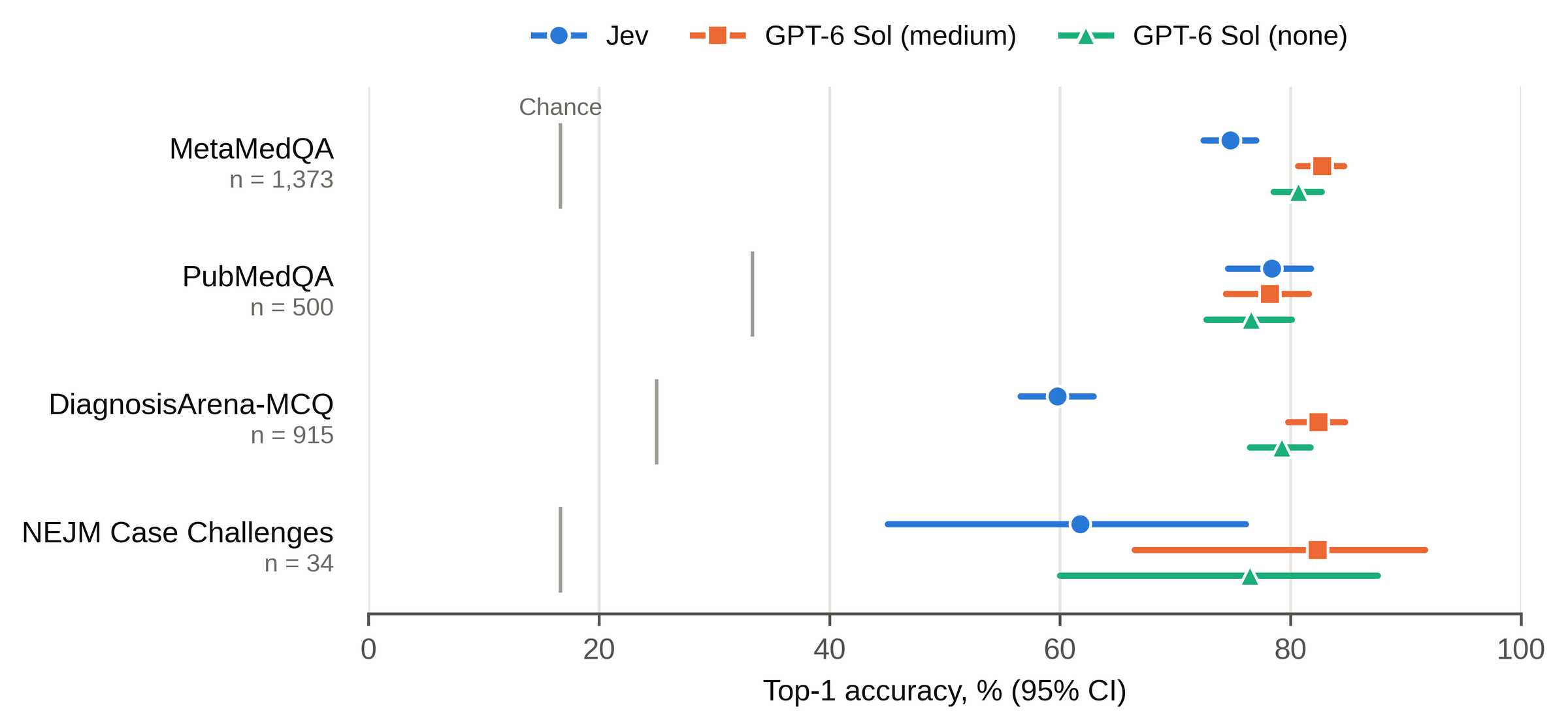}
\caption{Top-1 accuracy of Jev and GPT-6 Sol in each benchmark, with medium reasoning and without reasoning (none). Markers show accuracy and horizontal bars show Wilson 95\% CIs. The values are given in Table~\ref{tab:2}. Grey vertical lines mark chance accuracy (1/number of options). Below each benchmark name, n is the number of scored items. NEJM accuracy refers to the 34 cases with a published final diagnosis.}\label{fig:2}
\end{figure}

Jev's accuracy was well above chance in every benchmark, but its gap to GPT-6 Sol widened as the tasks moved from research abstracts to examination questions and complex diagnostic cases (Table~\ref{tab:2}; Figure~\ref{fig:2}). On PubMedQA, Jev's accuracy (78.4\%) was similar to that of GPT-6 Sol with medium reasoning (78.2\%; difference 0.2 percentage points, 95\% CI $-$2.2 to 2.6) and to the single-annotator accuracy reported by the dataset authors (78.0\%). On MetaMedQA, Jev was 7.9 percentage points less accurate (74.8\% vs 82.7\%). On the diagnostic benchmarks, the difference reached 22.6 points on DiagnosisArena-MCQ (59.8\% vs 82.4\%) and 20.6 points on the NEJM cases (61.8\% vs 82.4\%). The differences on MetaMedQA and DiagnosisArena-MCQ remained significant after Holm adjustment, but that on the NEJM cases did not (P = 0.078). In DiagnosisArena-MCQ, 237 cases were answered correctly by GPT-6 Sol alone and 30 by Jev alone (Supplementary Table~\ref{tab:S2}).

\begin{table}[tb]
\caption{Accuracy of Jev and GPT-6 Sol in each benchmark (primary outcome).}\label{tab:2}
\centering\small
\setlength{\tabcolsep}{5pt}
\begin{tabular}{@{}lcccccc@{}}
\toprule
\makecell[lb]{Benchmark\\(items; chance)} & Jev & \makecell[b]{GPT-6 Sol\\(medium)} & \makecell[b]{GPT-6 Sol\\(none)} & \makecell[b]{Jev $-$ GPT-6 Sol\\(medium), pp\\(95\% CI)} & \makecell[b]{P\\(Holm)} & \makecell[b]{Jev $-$ GPT-6 Sol\\(none), pp\\(95\% CI)}\\
\midrule
\makecell[l]{MetaMedQA\\(1,373; 16.7\%)} & \makecell[c]{74.8\\(72.4--77.0)} & \makecell[c]{82.7\\(80.6--84.6)} & \makecell[c]{80.7\\(78.5--82.7)} & \makecell[c]{$-$7.9\\($-$10.0 to $-$6.0)} & $<$0.001 & \makecell[c]{$-$5.9\\($-$7.8 to $-$4.2)}\\
\addlinespace[4pt]
\makecell[l]{PubMedQA\\(500; 33.3\%)} & \makecell[c]{78.4\\(74.6--81.8)} & \makecell[c]{78.2\\(74.4--81.6)} & \makecell[c]{76.6\\(72.7--80.1)} & \makecell[c]{0.2\\($-$2.2 to 2.6)} & 1.00 & \makecell[c]{1.8\\($-$1.0 to 4.8)}\\
\addlinespace[4pt]
\makecell[l]{DiagnosisArena-MCQ\\(915; 25.0\%)} & \makecell[c]{59.8\\(56.6--62.9)} & \makecell[c]{82.4\\(79.8--84.7)} & \makecell[c]{79.2\\(76.5--81.7)} & \makecell[c]{$-$22.6\\($-$25.9 to $-$19.7)} & $<$0.001 & \makecell[c]{$-$19.5\\($-$22.6 to $-$16.3)}\\
\addlinespace[4pt]
\makecell[l]{NEJM Case Challenges\\(34; 16.7\%)} & \makecell[c]{61.8\\(45.0--76.1)} & \makecell[c]{82.4\\(66.5--91.7)} & \makecell[c]{76.5\\(60.0--87.6)} & \makecell[c]{$-$20.6\\($-$35.3 to $-$5.9)} & 0.078 & \makecell[c]{$-$14.7\\($-$29.4 to 0.0)}\\
\bottomrule
\end{tabular}
\tabnote{Accuracy, \% (Wilson 95\% CI); no item lacked a valid response. Differences are paired, in percentage points (pp), with 95\% CIs from 2,000 bootstrap resamples of matched items; negative values favour GPT-6 Sol. P values are from the exact McNemar test for the main contrast (Jev vs GPT-6 Sol with medium reasoning), Holm-adjusted across the four benchmarks (unadjusted: \textless0.001, 1.00, \textless0.001 and 0.039). Unadjusted P values for Jev vs GPT-6 Sol without reasoning: \textless0.001, 0.27, \textless0.001 and 0.13. Chance, 1/number of options. NEJM Case Challenges, 34 cases with a published final diagnosis; in the 33 cases with a closed poll, the mean proportion of reader votes for the correct option was 31.2\%. Agreement between models, differences in Brier score and the comparison between GPT-6 Sol conditions are given in Supplementary Table~\ref{tab:S2}.}
\end{table}

Explicit reasoning explained little of this gap. Without reasoning, GPT-6 Sol lost 2.0 percentage points on MetaMedQA (95\% CI $-$3.1 to $-$0.9) and 3.2 on DiagnosisArena-MCQ (95\% CI $-$4.9 to $-$1.4) relative to medium reasoning, and it still exceeded Jev by 5.9 and 19.5 points, respectively (Table~\ref{tab:2}).

On PubMedQA, both models seldom recognised inconclusive evidence. Jev's recall of maybe was 27.3\% (15/55), compared with 85.1\% for yes and 84.0\% for no, so its macro-F1 (0.654) was below the single-annotator value (0.722); GPT-6 Sol with medium reasoning showed the same pattern (recall of maybe, 25.5\%). Jev's accuracy on DiagnosisArena-MCQ was close to the 61.9\% that the benchmark's authors reported for o1 on the full set of 1,113 cases, of which 915 are publicly available~\citep{Zhu2026}. In the 33 polled NEJM cases, the proportion of reader votes for the correct option averaged 31.2\%; Jev exceeded it by 32.5 percentage points (95\% CI 18.6 to 45.7) and GPT-6 Sol with medium reasoning by 50.6 (95\% CI 38.9 to 61.5) (Supplementary Table~\ref{tab:S3}).

\subsection{Calibration and selective prediction (Q2)}\label{sec:calib}

\begin{figure}[tbp]
\centering
\includegraphics[width=\linewidth]{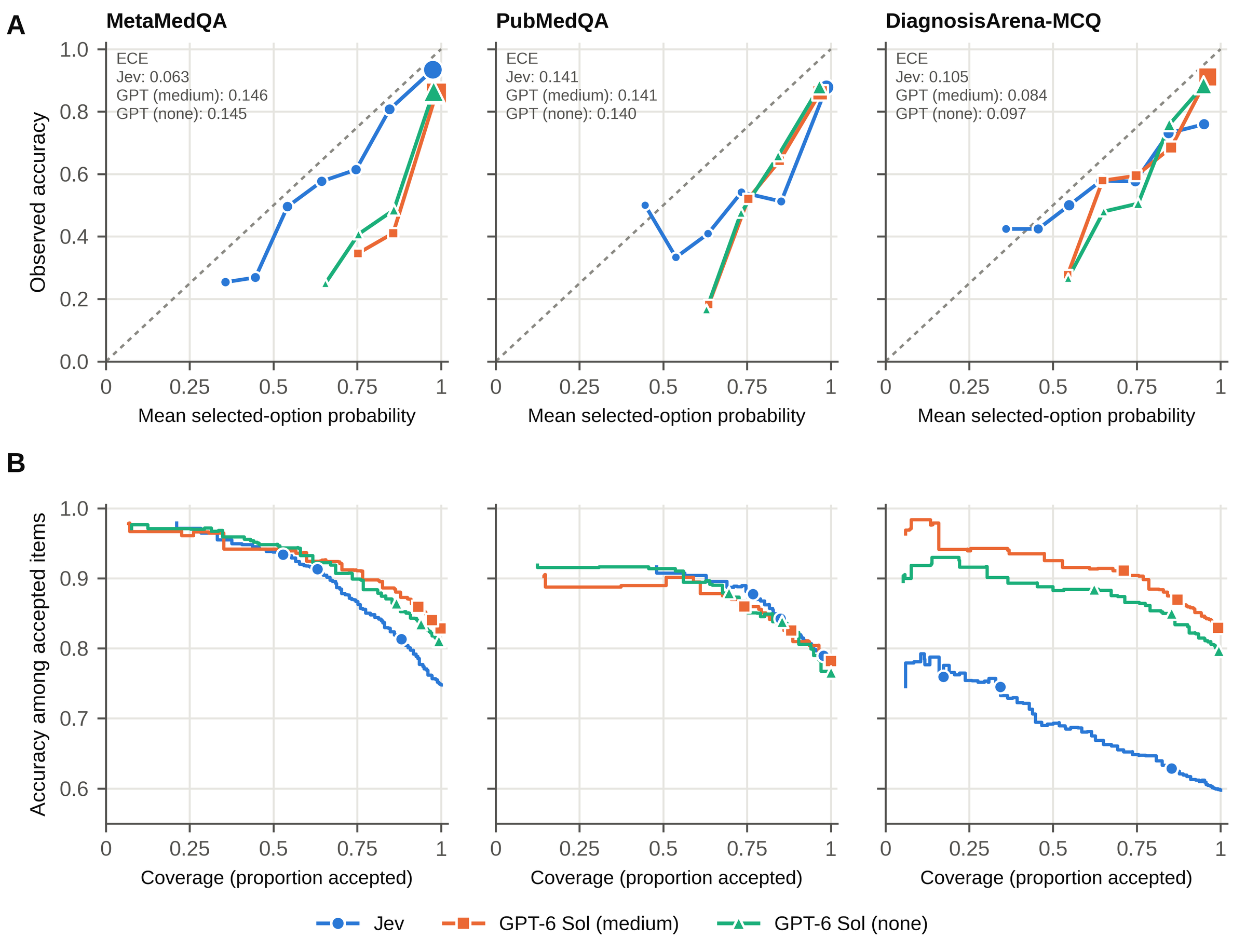}
\caption{Calibration (A) and selective prediction (B) in MetaMedQA, PubMedQA and DiagnosisArena-MCQ. (A) Reliability diagrams. Answers were grouped into ten equal-width bins of the selected-option probability. Each marker shows the mean probability and observed accuracy of one bin, and marker area increases with the number of answers in the bin. Bins with fewer than ten answers are not drawn but still count toward the expected calibration error (ECE) given in each panel. Points below the dashed diagonal indicate overconfidence. (B) Accuracy-coverage curves. An answer was accepted when its selected-option probability was at or above a threshold, and the threshold took every observed probability value. The curves are step functions, shown for coverages of at least 5\%. Markers indicate thresholds of 0.9, 0.8 and 0.5, from left to right. Probabilities are native for Jev and verbalised for GPT-6 Sol. The NEJM Case Challenges are not shown (Section~\ref{sec:stats}).}\label{fig:3}
\end{figure}

All models were overconfident on average: the mean probability of the selected option exceeded accuracy for every model in every benchmark, by 5.8 to 13.6 percentage points for Jev and by 8.4 to 16.0 for GPT-6 Sol (Table~\ref{tab:3}; Supplementary Table~\ref{tab:S3}). The models differed, however, in how they distributed their probabilities. GPT-6 Sol with medium reasoning assigned a probability of at least 0.9 to 93.2\% of its MetaMedQA answers and to 71.1\% of its DiagnosisArena-MCQ answers, whereas Jev did so for 52.9\% and 17.3\% (Figure~\ref{fig:3}).

\begin{table}[tbp]
\caption{Probability quality, calibration and selective prediction.}\label{tab:3}
\centering\small
\setlength{\tabcolsep}{5pt}
\begin{tabular}{@{}lccccccc@{}}
\toprule
Benchmark and model & \makecell[b]{Mean selected-\\option probability} & ECE & AUROC & \makecell[b]{Brier\\score} & \makecell[b]{Coverage at\\$\geq$0.9, \%} & \makecell[b]{Accuracy at $\geq$0.9,\\\% (95\% CI)} & AURC\\
\midrule
\multicolumn{8}{@{}l}{\textbf{MetaMedQA (n = 1,373)}}\\
\quad Jev & 0.811 & 0.063 & 0.845 & 0.343 & 52.9 & 93.4 (91.3--95.0) & 0.087\\
\quad GPT-6 Sol (medium) & 0.973 & 0.146 & 0.801 & 0.316 & 93.2 & 85.9 (83.9--87.7) & 0.070\\
\quad GPT-6 Sol (none) & 0.951 & 0.145 & 0.818 & 0.341 & 86.7 & 86.5 (84.4--88.3) & 0.071\\
\addlinespace[4pt]
\multicolumn{8}{@{}l}{\textbf{PubMedQA (n = 500)}}\\
\quad Jev & 0.920 & 0.141 & 0.766 & 0.350 & 76.8 & 87.8 (84.1--90.7) & 0.109\\
\quad GPT-6 Sol (medium) & 0.923 & 0.141 & 0.732 & 0.360 & 74.2 & 86.0 (82.1--89.1) & 0.126\\
\quad GPT-6 Sol (none) & 0.906 & 0.140 & 0.786 & 0.359 & 69.6 & 87.9 (84.1--90.9) & 0.114\\
\addlinespace[4pt]
\multicolumn{8}{@{}l}{\textbf{DiagnosisArena-MCQ (n = 915)}}\\
\quad Jev & 0.699 & 0.105 & 0.645 & 0.570 & 17.3 & 75.9 (68.7--81.9) & 0.303\\
\quad GPT-6 Sol (medium) & 0.908 & 0.084 & 0.768 & 0.293 & 71.1 & 91.1 (88.7--93.0) & 0.080\\
\quad GPT-6 Sol (none) & 0.889 & 0.097 & 0.728 & 0.344 & 62.3 & 88.4 (85.5--90.8) & 0.120\\
\bottomrule
\end{tabular}
\tabnote{Probabilities native for Jev and verbalised for GPT-6 Sol, rescaled to sum to 1. The Brier score uses the complete probability vector; the other columns use the probability of the selected option. Coverage is the proportion of all items whose selected-option probability was at or above 0.9, and accuracy refers to those items (Wilson 95\% CI); results at thresholds of 0.5 and 0.8 are given in Supplementary Table~\ref{tab:S4}. ECE, expected calibration error (ten equal-width bins; lower is better); AUROC, area under the receiver operating characteristic curve for discriminating correct from incorrect answers (higher is better); Brier score, multiclass (range 0--2; lower is better; not comparable across benchmarks because it depends on the number of options); AURC, area under the risk--coverage curve (lower is better). Paired differences with bootstrap 95\% CIs are reported in the text and in Supplementary Table~\ref{tab:S2}. The NEJM Case Challenges are not included (Section~\ref{sec:stats}); their mean selected-option probability and Brier score are given in Supplementary Table~\ref{tab:S3}.}
\end{table}

On MetaMedQA, Jev's probabilities were the best calibrated (ECE 0.063 vs 0.146 with medium reasoning; difference $-$0.083, 95\% CI $-$0.100 to $-$0.064) and discriminated better (AUROC 0.845 vs 0.801; difference 0.044, 95\% CI 0.014 to 0.074). Its answers with a probability of at least 0.9 were correct in 93.4\% of cases, whereas GPT-6 Sol was markedly overconfident below 0.9: with medium reasoning, its answers with probabilities from 0.8 to 0.9 were correct in only 41.1\%. Accordingly, only 48 of Jev's 346 errors (13.9\%) were made with a probability of at least 0.9, compared with 180 of GPT-6 Sol's 237 (75.9\%). This did not, however, translate into more answers that could be accepted at a given accuracy. At similar coverage, GPT-6 Sol was as accurate as Jev or more so (for example, 93.9\% for Jev at a coverage of 49.7\%, and 94.2\% and 94.8\% for GPT-6 Sol with and without reasoning at 52.8\% and 50.9\%), and across all thresholds the AURC was larger, i.e., worse, for Jev (0.087 vs 0.070; difference 0.017, 95\% CI 0.005 to 0.029).

On DiagnosisArena-MCQ, Jev's probabilities separated correct from incorrect answers poorly (AUROC 0.645 vs 0.768; difference $-$0.123, 95\% CI $-$0.172 to $-$0.077), although its ECE did not differ significantly from that of GPT-6 Sol (0.105 vs 0.084; difference 0.021, 95\% CI $-$0.007 to 0.055). Raising the threshold therefore improved its accuracy only modestly, to 74.5\% at 0.8 (coverage 34.3\%) and 75.9\% at 0.9 (coverage 17.3\%; Supplementary Table~\ref{tab:S4}), whereas GPT-6 Sol reached 91.1\% at 0.9 while accepting 71.1\% of cases (AURC 0.303 for Jev vs 0.080).

On PubMedQA, the three models behaved similarly (ECE 0.140--0.141; at a threshold of 0.9, coverage 69.6--76.8\% and accuracy 86.0--87.9\%), with a slightly smaller AURC for Jev than for GPT-6 Sol with medium reasoning (0.109 vs 0.126; difference $-$0.017, 95\% CI $-$0.035 to $-$0.001).

The ECE estimates were insensitive to the binning scheme: other schemes changed the ECE of each model by at most 0.005 (Supplementary~\ref{sec:S4}). Jev's confidence value discriminated correct from incorrect answers as well as its probability did (AUROC 0.844, 0.778 and 0.645 on MetaMedQA, PubMedQA and DiagnosisArena-MCQ), and the same thresholds applied to it gave lower coverage and higher accuracy (for example, on MetaMedQA at 0.9, coverage 49.8\% and accuracy 94.0\%; Supplementary Table~\ref{tab:S5}). In the NEJM cases, in which calibration was not assessed, Jev's mean selected-option probability was 0.676 and its Brier score 0.577, compared with 0.944 and 0.285 for GPT-6 Sol with medium reasoning.

\subsection{Recognition of unanswerable questions (Q3)}\label{sec:unans}

Jev answered 86.5\% (948/1,096) of the MetaMedQA questions with a correct substantive option correctly, and it selected ``None of the above'' for 53.9\% (62/115) of the questions for which that option was correct. It selected ``I don't know or cannot answer'', however, for only 10.5\% (17/162; 95\% CI 6.7--16.2) of the questions for which that was the correct answer, and chose a substantive option for 104 of them (64.2\%). The pattern was clearest for the 100 questions about a fictional organ, all of which were unanswerable: Jev selected a substantive option 52 times, ``None of the above'' 38 times and ``I don't know'' 10 times. When Jev did select ``I don't know'', it was almost always right (17 of 18 selections).

GPT-6 Sol was more accurate than Jev on answerable questions (94.6\% with medium reasoning) and had a higher missing-answer recall (73.9\%), but not a higher unknown recall (8.6\% with medium reasoning and 4.3\% without). Neither model had been told that questions might be unanswerable; for context, GPT-4o reached an unknown recall of 3.7\% with the baseline prompt of Griot et al.\ and 44.4--51.8\% with prompts that warned of this possibility. Cross-tabulations of reference and selected categories are given in Supplementary Table~\ref{tab:S6}.

\subsection{Response time and cost}\label{sec:cost}

Jev's median client-side latency was 0.27--0.31 s in every benchmark (95th percentile, 0.35--0.38 s), including the NEJM cases, whose requests contained a median of 2,364 input tokens. The median latency of GPT-6 Sol was 1.68--3.82 s with medium reasoning and 1.19--1.62 s without reasoning. Evaluating all 2,823 items cost US\$0.082 with Jev, US\$6.87 with GPT-6 Sol with medium reasoning and US\$3.94 without reasoning (Supplementary Table~\ref{tab:S7}).

\section{Discussion}\label{sec:discussion}

\subsection{Principal findings}\label{sec:principal}

In this evaluation of Jev on medical question-answering and case-based diagnostic-reasoning benchmarks, the model answered every item with one of the predefined options, with a median latency of 0.31 s or less and at a total cost below US\$0.10. On research questions about biomedical abstracts, its accuracy was similar to that of a frontier reasoning LLM and of a human annotator, but it fell behind by 8 percentage points on examination questions and by 21 to 23 points on complex diagnostic cases. Its better calibration on examination questions did not yield more answers that could be accepted at a given accuracy, its probabilities barely distinguished correct from incorrect answers on complex cases, and, like the generative models evaluated with MetaMedQA, it seldom chose ``I don't know or cannot answer'' when that was the correct answer.

\subsection{Interpretation}\label{sec:interpretation}

The accuracy gap is unlikely to be explained by explicit reasoning alone. Without reasoning tokens, GPT-6 Sol exceeded Jev by 19.5 percentage points on DiagnosisArena-MCQ and was only 1.6 to 5.9 points less accurate than with medium reasoning across the four benchmarks. The gap may therefore reflect differences in the medical knowledge encoded by the models or in their ability to integrate long, multi-part case descriptions, which our design cannot separate. The condition without reasoning tokens still generated its structured answer autoregressively and did not match Jev's inference architecture. The weaknesses with numerical content reported by the developer~\citep{TypeSafeAI2026jagged} may contribute, because clinical cases are rich in laboratory values, but we could not quantify their effect. In the 34 scored NEJM cases, Jev was not less accurate when laboratory tables were present (16 of 25 cases, vs 5 of 9 without), although these numbers are too small to exclude an effect. The construction of DiagnosisArena-MCQ, whose cases were filtered against LLMs and whose distractors derive from errors of reasoning models, could also have affected the comparison in either direction.

On MetaMedQA, Jev's probabilities were better calibrated than those of GPT-6 Sol, but Jev remained overconfident on average in every benchmark, so its probabilities should not be used as thresholds without local validation. The number of answers that can be accepted at a target accuracy, moreover, depends both on how accurate a model is and on how well its scores rank its own answers. Once its threshold has been set on local data, a more accurate model whose scores rank its answers adequately can accept as many answers at the same accuracy, even if it is miscalibrated; this is what we observed on MetaMedQA. Where ranking failed, as on DiagnosisArena-MCQ, no threshold compensated: even the most confident sixth of Jev's answers (probability $\geq$0.9) were wrong about once in four. Jev's calibration also varied across tasks (ECE from 0.063 on MetaMedQA to 0.141 on PubMedQA), in line with the recommendation of a previous evaluation of Jev in radiology to recalibrate its probabilities for each task~\citep{Huang2026}.

Restricting outputs to predefined options removes one form of hallucination, the fabrication of non-existent answers, but not the confident selection of a wrong or meaningless one. Faced with questions about an organ that does not exist, Jev chose a substantive answer or ``None of the above'' 90 times out of 100. Abstention may therefore need to be designed into the task rather than expected from the model. One option is to tell the model that some questions may be unanswerable, which substantially increased the unknown recall of generative models on MetaMedQA; another is to ask a separate question about whether the information provided is sufficient. Neither approach was tested here. Unknown recall also measures explicit abstention only; we did not evaluate whether Jev's low probabilities, which its developer presents as the model's built-in signal of uncertainty, could flag unanswerable questions through threshold-based deferral.

These findings support evaluating Jev as a component of clinical workflows, not as a decision-maker. Its speed and cost make it attractive for high-volume, text-based decisions among predefined options, such as screening articles for systematic reviews, checking trial eligibility criteria or verifying the outputs of generative models with uncertain answers deferred to clinicians. Our results suggest that such uses are most promising when the answer depends on reading a single text, as in PubMedQA, and least promising for multi-step diagnostic reasoning. It remains to be tested whether decomposing complex judgements into atomic questions combined by explicit rules, or converting numerical data into named categories before the question is asked narrows the gap on complex cases.

\subsection{Limitations}\label{sec:limitations}

First, Jev accepts text input only, so our results do not extend to decisions that depend on images or signals. In the NEJM cases, figures were replaced by their captions, whereas readers could see them, and some malformed MetaMedQA questions refer to media that are not available. Numerical content, a documented weakness of the model, was presented unchanged so that the results reflect the original benchmarks. Moreover, every diagnostic item required choosing among supplied options, so the results reflect recognition of the correct diagnosis among alternatives rather than the generation of a differential diagnosis. In DiagnosisArena, for example, o1 answered 31.1\% of the cases correctly in the open-ended format and 61.9\% in the multiple-choice format.

Second, data contamination cannot be excluded. The public benchmarks and their answers were published before the release of both models, the final diagnoses of the NEJM cases are available online, and the training data of neither model have been disclosed; exposure could inflate accuracy and bias the comparison in either direction. The benchmarks may also overlap, as DiagnosisArena includes NEJM case reports, and reference answers were used as published, without clinical adjudication. An evaluation on items released after the models' training would provide an estimate free of contamination.

Third, each model and condition was run once, with options in their original order, so repeatability and sensitivity to option order were not assessed.

GPT-6 Sol was evaluated at two of its six reasoning-effort levels, with a single unoptimised prompt, and its probabilities were verbalised rather than native. Because the verbalised confidence of LLMs is often poorly calibrated, differences in probability quality partly reflect how the probabilities were obtained. Jev was accessed through an intermediary, which may have added latency. The developer reports that most requests complete in about 100 ms, whereas our median client-side latency was 0.27--0.31 s. Finally, the study was not designed to establish equivalence, so the similar accuracy on PubMedQA should not be read as such. The comparison with NEJM readers is descriptive: readers were self-selected and of unknown expertise, only 33 cases were available, and the published poll percentages summed to 96--101\%.

\section{Conclusion}\label{sec:conclusion}

Jev was fast and inexpensive, and its accuracy was similar to that of a frontier reasoning LLM on questions about research abstracts, but lower on examination questions and much lower on complex diagnostic cases. Its better calibration on examination questions gave no selective-prediction advantage, its probabilities discriminated poorly on complex cases, and it rarely chose ``I don't know'' when that was the correct answer. Jev could therefore be explored as a low-cost component for text-based decisions among predefined options, with uncertain answers deferred to clinicians, but not as a stand-alone diagnostic tool. Before any clinical use, its accuracy and calibration should be verified for each intended task, and its probabilities recalibrated if necessary.

\section*{Declarations}

\subsection*{Author contributions (CRediT)}

\textbf{Alfredo Madrid-García}: Conceptualization, Writing -- original draft, Writing -- review \& editing, Visualization, Software, Supervision, Resources

\textbf{Beatriz Merino-Barbancho:} Conceptualization, Writing -- review \& editing.

All authors have seen and approved the final manuscript.

\subsection*{Funding}

This article did not receive any specific grant from funding agencies in the public, commercial, or not-for-profit sectors.

\subsection*{Conflict of interest}

AMG is employed as a data scientist by Roche. This work was carried out independently of that employment, without the use of company facilities, tools, data or confidential information. The views expressed herein are solely those of the author and do not represent the views or official position of Roche.

BMB declares no competing interests.

\subsection*{Data availability}

This study analysed existing benchmarks and generated new model outputs. Code, dataset revisions, sample identifiers, preprocessing rules and shareable model outputs are freely available and can be downloaded as supplementary data from \url{https://zenodo.org/records/23008156}. Copyrighted NEJM case texts are not redistributed; case identifiers and derived results are provided.

\subsection*{Ethics approval}

This study used publicly available examination questions, research abstracts and previously published, de-identified clinical case descriptions. No patients were recruited and no clinical records were accessed. For this reason, ethics committee approval was not required.

\subsection*{Generative AI statement}

Generative AI models, specifically Anthropic's Claude Opus 5.5 (high) and OpenAI's GPT-6 Astra (high) were used extensively to support the research workflow, data analysis, manuscript drafting, and editing.

\subsection*{Patient consent for publication}

Not applicable.

\subsection*{Acknowledgements}

Not applicable.

\bibliographystyle{unsrtnat}
\bibliography{references}

\clearpage
\setcounter{section}{0}
\setcounter{table}{0}
\setcounter{figure}{0}
\renewcommand{\thesection}{S\arabic{section}}
\renewcommand{\thetable}{S\arabic{table}}
\renewcommand{\thefigure}{S\arabic{figure}}
\renewcommand{\theHsection}{S\arabic{section}}
\renewcommand{\theHtable}{S\arabic{table}}
\renewcommand{\theHfigure}{S\arabic{figure}}

\begin{center}
  {\Large\bfseries Supplementary Material}\\[6pt]
  {\large Jev in Medicine: A Benchmark Evaluation}\\[4pt]
  \emph{Alfredo Madrid-García, Beatriz Merino-Barbancho}
\end{center}

\paragraph{Contents}\mbox{}
\begin{itemize}
  \item[] \ref{sec:S1}. Datasets
  \item[] \ref{sec:S2}. Prompts and requests (Table~\ref{tab:S1})
  \item[] \ref{sec:S3}. Statistical methods
  \item[] \ref{sec:S4}. Additional results (Tables~\ref{tab:S2}--\ref{tab:S7})
\end{itemize}

\section{Datasets}\label{sec:S1}

\textbf{MetaMedQA.} Test split of the Hugging Face dataset maximegmd/MetaMedQA. As in the original benchmark code, each item was identified by its zero-based row index. The counts by reference category (1,096 A--D, 115 E and 162 F) and by examination label (Step 1, 679; Step 2\&3, 594; fictional organ, 100) were verified. Row 1198 is labelled as an original question but has option F recorded as ``AR''; it was kept unchanged.

\textbf{PubMedQA.} Test split generated from the official repository with the unmodified split script, which reproduced the distributed ground-truth file in identifiers, labels and order, 500 instances (276 yes, 169 no and 55 maybe), identified by their PubMed identifier. The input contained the abstract sections other than the conclusions, each preceded by its original heading.

\textbf{DiagnosisArena-MCQ.} Test split of the Hugging Face dataset SII-SPIRAL-MED/DiagnosisArena, with the case number as item identifier. The input contained the fields Case Information, Physical Examination and Diagnostic Tests.

\textbf{NEJM Case Challenges.} Extracted from nejm.org on 24 September 2026 and numbered from the most recent (case 1) to the oldest (case 35). Case 1 (23 September 2026) had no published final diagnosis, and the polls for cases 1 and 2 were still open. Readers submitted 273,362 responses to the 33 closed polls (median 7,732 per case; range 4,386--17,714). The percentages in each poll summed to 96--101\% and were used without adjustment. The correct option of each case was identified by hand, by matching the final diagnosis published in the Case Record to the poll options. Case texts are protected by copyright and are not redistributed.

\section{Prompts and requests}\label{sec:S2}

Both models received the same instruction, case content and answer options (Table~\ref{tab:S1}). No system prompt, worked examples or task-specific guidance were used, except for an output-format instruction for GPT-6 Sol. Prompts were not optimised. Reference answers, reader votes and final diagnoses were never added to the inputs, and text already present in the items was kept unchanged.

\begin{table}[tbp]
\caption{Instruction, case content and answer options for each dataset.}\label{tab:S1}
\centering\small
\begin{tabularx}{\linewidth}{@{}>{\raggedright\arraybackslash}p{2.3cm}LLL@{}}
\toprule
Dataset & Instruction & Case content (Jev state) & Answer options\\
\midrule
MetaMedQA & ``Which option is the correct answer to this question?'' & Complete original question, including its final question sentence (string) & A--F with their original text and order; E, ``None of the above''; F, ``I don't know or cannot answer''\\
\addlinespace[4pt]
PubMedQA & The research question, verbatim (not repeated elsewhere) & Abstract without its conclusions, each section as ``HEADING: text'' (object with one field, Abstract) & yes, no and maybe, each with a one-sentence description\\
\addlinespace[4pt]
DiagnosisArena-MCQ & ``What is the most likely diagnosis for this patient?'' & Case Information, Physical Examination and Diagnostic Tests (object with these three fields) & A--D with their original text and order\\
\addlinespace[4pt]
NEJM Case Challenges & The poll question, verbatim (``What is the most likely diagnosis in this case?''; case 12, ``The most likely diagnosis is:'') & Case presentation up to the diagnostic question, with laboratory tables as text and figures replaced by their captions (string) & A--F with their original text, in the published order\\
\bottomrule
\end{tabularx}
\tabnote{GPT-6 Sol received the same three elements in a single user message.}
\end{table}

\textbf{Jev.} Each item was sent as one request to the System One endpoint, with a single Choice question:

\begin{Verbatim}[fontsize=\small,frame=leftline,xleftmargin=1em,framesep=2mm]
{"model": "typesafe/jev-1.13",
"state": <case content>,
"questions": {"answer": {"type": "choice",
"instructions": <instruction>,
"criteria": {"<option key>": "<option text>", ...}}}}
\end{Verbatim}

The question key (``answer'', or ``diagnosis'' for DiagnosisArena-MCQ and the NEJM Case Challenges) only labels the answer and is not passed to the model. For PubMedQA, the criteria were \emph{yes}, ``The evidence reported in the abstract supports a positive answer to the research question.''; \emph{no}, ``The evidence reported in the abstract supports a negative answer to the research question.''; and \emph{maybe}, ``The evidence reported in the abstract is inconclusive or mixed, so it supports neither a definite yes nor a definite no.''

\textbf{GPT-6 Sol.} Each item was sent as a single user message built from the same template:

\begin{Verbatim}[fontsize=\small,frame=leftline,xleftmargin=1em,framesep=2mm]
{instruction}

<Section label>:
{case content}

Options:
{one line per option: "A. option text", or "yes: description" for PubMedQA}

Return the letter of the correct answer in "answer" and, in "probabilities", your probability
(between 0 and 1) that each option is the correct answer. The six probabilities must sum to 1.
\end{Verbatim}

The section labels were ``Question:'' (MetaMedQA), ``Abstract:'' (PubMedQA), ``Case Information:'', ``Physical Examination:'' and ``Diagnostic Tests:'' (DiagnosisArena-MCQ) and ``Case presentation:'' (NEJM Case Challenges). The final sentence named the number of options and, for the diagnostic benchmarks, asked for ``the letter of the most likely diagnosis'' and the probability that each option ``is the correct diagnosis''; for PubMedQA, it asked for ``yes'', ``no'' or ``maybe''. The output was constrained with a strict JSON schema that admitted only the available options as the answer and required one number per option in ``probabilities'', with no additional properties. The complete request of every item was logged.

\section{Statistical methods}\label{sec:S3}

\textbf{Accuracy and paired comparisons.} Accuracy was computed over all scored items, with items lacking a valid response counted as incorrect, and reported with Wilson 95\% CIs; for PubMedQA, macro-F1 was the unweighted mean of the F1 scores of the three labels. Accuracy was scored from the answer returned by the API, without replacing it with the highest-probability option. In three responses the returned answer was not the highest-probability option: one from Jev in MetaMedQA (scored as correct; the highest-probability option was incorrect), one from Jev in DiagnosisArena-MCQ (scored as incorrect; the highest-probability option was correct) and one from GPT-6 Sol without reasoning in MetaMedQA (incorrect under either rule). Paired differences in accuracy and Brier score were estimated with a percentile bootstrap (2,000 resamples of matched items), accuracy was also compared with the exact two-sided McNemar test, and agreement on the selected option was summarised with Cohen's kappa. For the main contrast (Jev vs GPT-6 Sol with medium reasoning), the four P values were Holm-adjusted.

\textbf{Probability-based measures.} All probability vectors were rescaled to sum to 1. Calibration, discrimination and selective-prediction measures used the probability of the selected option. The expected calibration error (ECE) used ten equal-width bins, omitting empty bins. The AUROC was computed from ranks, with average ranks for ties and correct answers as the positive class. The Brier score used the complete vector, the sum, over all options, of the squared difference between the probability of each option and 1 for the correct option or 0 otherwise (range 0--2), averaged over items. For selective prediction, coverage was computed over all scored items and accuracy over the accepted answers at thresholds of 0.5, 0.8 and 0.9; in MetaMedQA, accepted ``None of the above'' and ``I don't know'' answers were also counted separately. The area under the risk--coverage curve (AURC) was the area under the step function of risk (1 $-$ accuracy of the accepted answers) against coverage over all observed probability values; a model whose probabilities did not rank its answers would have an AURC close to its error rate. The ECE, AUROC, selective prediction and AURC were not computed for the NEJM Case Challenges; their mean selected-option probability and Brier score are reported in Tables~\ref{tab:S2} and~\ref{tab:S3}.

\textbf{Additional analyses.} 95\% CIs for the paired differences in ECE, AUROC and AURC were obtained from 2,000 bootstrap resamples (percentile method; the same resampled items for the three models). The ECE was recomputed with 15 and 20 equal-width bins and with ten bins of equal size, and the comparison with readers was repeated after rescaling each poll to 100\%. For Jev, the same answers were also scored with the confidence value instead of the selected-option probability (mean, AUROC, AURC, and coverage and accuracy at thresholds of 0.5, 0.8 and 0.9; Table~\ref{tab:S5}), with 95\% CIs for the differences from the same 2,000 bootstrap resamples, and the agreement between the two scores was summarised with the Spearman correlation. The confidence value was treated as a score for ranking and accepting answers, not as a probability of correctness, so no ECE or Brier score was computed for it; for the NEJM Case Challenges, only the means were compared.

\textbf{Comparison with NEJM readers.} Reader accuracy in each case was the proportion of votes for the correct option. The 95\% CI of the mean paired difference between model correctness and reader accuracy was obtained from 10,000 case-bootstrap resamples. We also simulated 10,000 reader profiles, each answering each case correctly with the observed proportion of correct votes, independently across cases, and report the percentage of simulated totals strictly below each model's number of correct answers, with ties reported separately.

\textbf{Token log-probabilities (exploratory).} In the none condition, GPT-6 Sol requests also asked for token log-probabilities with five alternatives, the maximum accepted by the API, read at the token that carried the answer; this token matched the answer in every response. The API returned fewer alternatives than requested (a single one for 758 of 915 DiagnosisArena-MCQ answers and 461 of 500 PubMedQA answers), which always covered at least 98\% of the probability. For DiagnosisArena-MCQ and PubMedQA, options absent from the returned alternatives were set to zero (an approximation that omits at most 0.02 of probability per answer); for MetaMedQA and the NEJM Case Challenges, all six options were never returned, so no distribution could be derived.

\section{Additional results}\label{sec:S4}

Tables~\ref{tab:S2}--\ref{tab:S7} complement Sections~\ref{sec:acc}--\ref{sec:cost} of the manuscript. Two sensitivity analyses, the exploratory log-probability analysis and the analysis of Jev's confidence value did not change any conclusion.

\textbf{Binning of the ECE.} With 15 or 20 equal-width bins or ten bins of equal size, the ECE of each model changed by at most 0.005, and the ordering of the models changed only in PubMedQA, where the three values differed by less than 0.006.

\textbf{Reader polls rescaled to 100\%.} Mean reader accuracy became 31.7\%, and the differences from readers changed by less than 0.6 percentage points (Table~\ref{tab:S3}).

\textbf{Token log-probabilities.} In the none condition, probabilities derived from token log-probabilities were more concentrated than the verbalised probabilities of the same answers (mean selected-option probability 0.964 vs 0.889 in DiagnosisArena-MCQ and 0.981 vs 0.906 in PubMedQA) and less informative, the ECE was 0.184 vs 0.097 and 0.228 vs 0.140, the AUROC 0.670 vs 0.728 and 0.560 vs 0.786, and the Brier score higher by 0.038 (95\% CI 0.022 to 0.053) and 0.100 (95\% CI 0.074 to 0.128).

\textbf{Confidence value of Jev.} The confidence value was lower than the selected-option probability in 2,306 of the 2,822 scored answers, equal to it in 515 and higher in one (mean difference $-$0.040 in MetaMedQA, $-$0.041 in PubMedQA, $-$0.101 in DiagnosisArena-MCQ and $-$0.068 in the NEJM Case Challenges), and the two scores ranked the answers almost identically (Spearman correlation 0.964--1.000). The AUROC and the AURC therefore differed by at most 0.011 and 0.008, with 95\% CIs that included zero, and each threshold accepted fewer answers with higher accuracy (at 0.9, coverage 49.8\%, 72.6\% and 13.6\% and accuracy 94.0\%, 89.0\% and 78.2\% in MetaMedQA, PubMedQA and DiagnosisArena-MCQ; Table~\ref{tab:S5}).

\begin{landscape}
\begin{table}[p]
\caption{Paired comparisons between Jev and GPT-6 Sol, and between the two GPT-6 Sol conditions.}\label{tab:S2}
\centering\small
\setlength{\tabcolsep}{5pt}
\begin{tabular}{@{}lccccccc@{}}
\toprule
\makecell[lb]{Comparison\\(first $-$ second)} & \makecell[b]{Accuracy, pp\\(95\% CI)} & \makecell[b]{Correct by first /\\second only; P} & \makecell[b]{Agreement,\\\% ($\kappa$)} & \makecell[b]{Brier score\\(95\% CI)} & \makecell[b]{ECE\\(95\% CI)} & \makecell[b]{AUROC\\(95\% CI)} & \makecell[b]{AURC\\(95\% CI)}\\
\midrule
\multicolumn{8}{@{}l}{\textbf{MetaMedQA (1,373 questions)}}\\
Jev $-$ GPT-6 Sol (medium) & \makecell[c]{$-$7.9\\($-$10.0 to $-$6.0)} & \makecell[c]{38 / 147;\\$<$0.001} & 83.0 (0.79) & \makecell[c]{0.027\\($-$0.001 to 0.055)} & \makecell[c]{$-$0.083\\($-$0.100 to $-$0.064)} & \makecell[c]{0.044\\(0.014 to 0.074)} & \makecell[c]{0.017\\(0.005 to 0.029)}\\
\addlinespace[3pt]
Jev $-$ GPT-6 Sol (none) & \makecell[c]{$-$5.9\\($-$7.8 to $-$4.2)} & \makecell[c]{44 / 125;\\$<$0.001} & 83.5 (0.79) & \makecell[c]{0.002\\($-$0.023 to 0.029)} & \makecell[c]{$-$0.081\\($-$0.098 to $-$0.062)} & \makecell[c]{0.027\\($-$0.001 to 0.053)} & \makecell[c]{0.016\\(0.006 to 0.027)}\\
\addlinespace[3pt]
\makecell[l]{GPT-6 Sol (none) $-$\\GPT-6 Sol (medium)} & \makecell[c]{$-$2.0\\($-$3.1 to $-$0.9)} & \makecell[c]{13 / 41;\\$<$0.001} & 93.4 (0.92) & \makecell[c]{0.025\\(0.008 to 0.043)} & \makecell[c]{$-$0.001\\($-$0.012 to 0.010)} & \makecell[c]{0.017\\($-$0.006 to 0.042)} & \makecell[c]{0.001\\($-$0.009 to 0.010)}\\
\addlinespace[3pt]
\addlinespace[4pt]
\multicolumn{8}{@{}l}{\textbf{PubMedQA (500 instances)}}\\
Jev $-$ GPT-6 Sol (medium) & \makecell[c]{0.2\\($-$2.2 to 2.6)} & \makecell[c]{18 / 17;\\1.00} & 91.4 (0.85) & \makecell[c]{$-$0.010\\($-$0.036 to 0.016)} & \makecell[c]{0.000\\($-$0.024 to 0.025)} & \makecell[c]{0.035\\($-$0.017 to 0.085)} & \makecell[c]{$-$0.017\\($-$0.035 to $-$0.001)}\\
\addlinespace[3pt]
Jev $-$ GPT-6 Sol (none) & \makecell[c]{1.8\\($-$1.0 to 4.8)} & \makecell[c]{31 / 22;\\0.27} & 88.2 (0.80) & \makecell[c]{$-$0.008\\($-$0.039 to 0.023)} & \makecell[c]{0.001\\($-$0.027 to 0.031)} & \makecell[c]{$-$0.020\\($-$0.068 to 0.025)} & \makecell[c]{$-$0.005\\($-$0.021 to 0.010)}\\
\addlinespace[3pt]
\makecell[l]{GPT-6 Sol (none) $-$\\GPT-6 Sol (medium)} & \makecell[c]{$-$1.6\\($-$4.0 to 0.8)} & \makecell[c]{14 / 22;\\0.24} & 91.6 (0.86) & \makecell[c]{$-$0.002\\($-$0.030 to 0.024)} & \makecell[c]{$-$0.001\\($-$0.023 to 0.022)} & \makecell[c]{0.055\\(0.016 to 0.097)} & \makecell[c]{$-$0.012\\($-$0.025 to 0.000)}\\
\addlinespace[3pt]
\addlinespace[4pt]
\multicolumn{8}{@{}l}{\textbf{DiagnosisArena-MCQ (915 cases)}}\\
Jev $-$ GPT-6 Sol (medium) & \makecell[c]{$-$22.6\\($-$25.9 to $-$19.7)} & \makecell[c]{30 / 237;\\$<$0.001} & 67.7 (0.57) & \makecell[c]{0.277\\(0.241 to 0.314)} & \makecell[c]{0.021\\($-$0.007 to 0.055)} & \makecell[c]{$-$0.123\\($-$0.172 to $-$0.077)} & \makecell[c]{0.222\\(0.186 to 0.259)}\\
\addlinespace[3pt]
Jev $-$ GPT-6 Sol (none) & \makecell[c]{$-$19.5\\($-$22.6 to $-$16.3)} & \makecell[c]{43 / 221;\\$<$0.001} & 68.3 (0.58) & \makecell[c]{0.226\\(0.186 to 0.266)} & \makecell[c]{0.009\\($-$0.018 to 0.042)} & \makecell[c]{$-$0.083\\($-$0.133 to $-$0.035)} & \makecell[c]{0.183\\(0.147 to 0.221)}\\
\addlinespace[3pt]
\makecell[l]{GPT-6 Sol (none) $-$\\GPT-6 Sol (medium)} & \makecell[c]{$-$3.2\\($-$4.9 to $-$1.4)} & \makecell[c]{22 / 51;\\$<$0.001} & 91.0 (0.88) & \makecell[c]{0.051\\(0.024 to 0.078)} & \makecell[c]{0.013\\($-$0.005 to 0.031)} & \makecell[c]{$-$0.040\\($-$0.080 to $-$0.002)} & \makecell[c]{0.039\\(0.021 to 0.060)}\\
\addlinespace[3pt]
\addlinespace[4pt]
\multicolumn{8}{@{}l}{\textbf{NEJM Case Challenges (34 cases)}}\\
Jev $-$ GPT-6 Sol (medium) & \makecell[c]{$-$20.6\\($-$35.3 to $-$5.9)} & \makecell[c]{1 / 8;\\0.039} & 67.6 (0.60) & \makecell[c]{0.292\\(0.089 to 0.502)} & --- & --- & ---\\
\addlinespace[3pt]
Jev $-$ GPT-6 Sol (none) & \makecell[c]{$-$14.7\\($-$29.4 to 0.0)} & \makecell[c]{1 / 6;\\0.13} & 67.6 (0.60) & \makecell[c]{0.194\\($-$0.012 to 0.396)} & --- & --- & ---\\
\addlinespace[3pt]
\makecell[l]{GPT-6 Sol (none) $-$\\GPT-6 Sol (medium)} & \makecell[c]{$-$5.9\\($-$14.7 to 0.0)} & \makecell[c]{0 / 2;\\0.50} & 88.2 (0.85) & \makecell[c]{0.098\\(0.013 to 0.219)} & --- & --- & ---\\
\bottomrule
\end{tabular}
\tabnote{Differences are first minus second, in percentage points (pp) for accuracy. Negative differences in accuracy and AUROC, and positive differences in Brier score, ECE and AURC, favour the second model or condition. 95\% CIs from 2,000 paired bootstrap resamples of items; P values from the exact two-sided McNemar test (unadjusted; Holm-adjusted P values for Jev vs GPT-6 Sol with medium reasoning are given in Table~\ref{tab:2} of the manuscript). ECE, AUROC and AURC use the selected-option probability; they were not computed for the NEJM Case Challenges. Differences in PubMedQA macro-F1 (95\% CI): Jev $-$ GPT-6 Sol (medium), 0.006 ($-$0.036 to 0.046); Jev $-$ GPT-6 Sol (none), $-$0.017 ($-$0.060 to 0.027); GPT-6 Sol (none) $-$ GPT-6 Sol (medium), 0.023 ($-$0.013 to 0.061). AUROC, area under the receiver operating characteristic curve; AURC, area under the risk--coverage curve; ECE, expected calibration error; $\kappa$, Cohen's kappa.}
\end{table}
\end{landscape}

\begin{table}[tbp]
\caption{NEJM Case Challenges. Accuracy of Jev and GPT-6 Sol and comparison with readers.}\label{tab:S3}
\centering\small
\setlength{\tabcolsep}{5pt}
\begin{tabular}{@{}>{\raggedright\arraybackslash}p{5.6cm}cccc@{}}
\toprule
Measure & Jev & \makecell[b]{GPT-6 Sol\\(medium)} & \makecell[b]{GPT-6 Sol\\(none)} & Readers\\
\midrule
\multicolumn{5}{@{}l}{\textbf{34 cases with a published final diagnosis}}\\
Accuracy, \% (n correct; 95\% CI) & \makecell[c]{61.8 (21;\\45.0--76.1)} & \makecell[c]{82.4 (28;\\66.5--91.7)} & \makecell[c]{76.5 (26;\\60.0--87.6)} & ---\\
\addlinespace[3pt]
Mean selected-option probability & 0.676 & 0.944 & 0.925 & ---\\
\addlinespace[3pt]
Brier score & 0.577 & 0.285 & 0.383 & ---\\
\addlinespace[3pt]
\addlinespace[4pt]
\multicolumn{5}{@{}l}{\textbf{33 cases with a final diagnosis and a closed poll (273,362 votes)}}\\
Accuracy, \% (n correct); readers, mean \% of votes for the correct option & 63.6 (21) & 81.8 (27) & 75.8 (25) & 31.2\\
\addlinespace[3pt]
Mean paired difference from readers, pp (95\% CI) & \makecell[c]{32.5\\(18.6 to 45.7)} & \makecell[c]{50.6\\(38.9 to 61.5)} & \makecell[c]{44.6\\(32.7 to 56.1)} & ---\\
\addlinespace[3pt]
Polls rescaled to 100\%: mean paired difference, pp (95\% CI); readers, mean \% & \makecell[c]{32.0\\(18.2 to 45.2)} & \makecell[c]{50.2\\(38.5 to 61.0)} & \makecell[c]{44.1\\(32.2 to 55.6)} & 31.7\\
\addlinespace[3pt]
Simulated reader profiles with fewer correct answers than the model, \% of 10,000 & 100 & 100 & 100 & ---\\
\addlinespace[3pt]
Selected the option most voted by readers, n (\%) & 19 (57.6) & 19 (57.6) & 18 (54.5) & \makecell[c]{Most-voted option\\correct: 16 (48.5)}\\
\addlinespace[3pt]
Cases with laboratory tables (24), \% (n correct) & 66.7 (16) & 79.2 (19) & 75.0 (18) & 31.8\\
\addlinespace[3pt]
Cases without laboratory tables (9), \% (n correct) & 55.6 (5) & 88.9 (8) & 77.8 (7) & 29.4\\
\bottomrule
\end{tabular}
\tabnote{Reader accuracy in each case is the proportion of votes for the correct option; values for readers are means across cases. The mean paired difference is the mean, across cases, of model correctness (0 or 1) minus reader accuracy (95\% CI from 10,000 case-bootstrap resamples). In the rescaled analysis, the published percentages of each poll, which summed to 96--101\%, were divided by their sum. Readers could see the figures, whereas the models received their captions only. The comparison with readers is descriptive. pp, percentage points.}
\end{table}

\begin{table}[tbp]
\caption{Selective prediction. Coverage and accuracy of accepted answers at thresholds of 0.5, 0.8 and 0.9.}\label{tab:S4}
\centering\small
\setlength{\tabcolsep}{5pt}
\begin{tabular}{@{}lcccccc@{}}
\toprule
Model & \makecell[b]{Coverage\\$\geq$0.5, \%} & \makecell[b]{Accuracy $\geq$0.5,\\\% (95\% CI)} & \makecell[b]{Coverage\\$\geq$0.8, \%} & \makecell[b]{Accuracy $\geq$0.8,\\\% (95\% CI)} & \makecell[b]{Coverage\\$\geq$0.9, \%} & \makecell[b]{Accuracy $\geq$0.9,\\\% (95\% CI)}\\
\midrule
\multicolumn{7}{@{}l}{\textbf{MetaMedQA (n = 1,373)}}\\
\quad Jev & 88.1 & 81.3 (79.0--83.4) & 63.1 & 91.3 (89.3--93.0) & 52.9 & 93.4 (91.3--95.0)\\
\quad GPT-6 Sol, medium & 99.9 & 82.9 (80.8--84.8) & 97.3 & 84.1 (82.0--85.9) & 93.2 & 85.9 (83.9--87.7)\\
\quad GPT-6 Sol, none & 99.3 & 81.1 (78.9--83.1) & 94.0 & 83.5 (81.4--85.4) & 86.7 & 86.5 (84.4--88.3)\\
\addlinespace[4pt]
\multicolumn{7}{@{}l}{\textbf{PubMedQA (n = 500)}}\\
\quad Jev & 97.8 & 78.9 (75.1--82.3) & 85.0 & 84.2 (80.5--87.4) & 76.8 & 87.8 (84.1--90.7)\\
\quad GPT-6 Sol, medium & 100.0 & 78.2 (74.4--81.6) & 88.2 & 82.5 (78.7--85.8) & 74.2 & 86.0 (82.1--89.1)\\
\quad GPT-6 Sol, none & 100.0 & 76.6 (72.7--80.1) & 85.4 & 83.8 (80.1--87.0) & 69.6 & 87.9 (84.1--90.9)\\
\addlinespace[4pt]
\multicolumn{7}{@{}l}{\textbf{DiagnosisArena-MCQ (n = 915)}}\\
\quad Jev & 85.4 & 62.9 (59.4--66.2) & 34.3 & 74.5 (69.4--79.0) & 17.3 & 75.9 (68.7--81.9)\\
\quad GPT-6 Sol, medium & 99.2 & 82.9 (80.3--85.2) & 87.1 & 87.0 (84.4--89.1) & 71.1 & 91.1 (88.7--93.0)\\
\quad GPT-6 Sol, none & 99.5 & 79.7 (76.9--82.2) & 85.4 & 85.0 (82.3--87.4) & 62.3 & 88.4 (85.5--90.8)\\
\bottomrule
\end{tabular}
\tabnote{Answers were accepted when the probability of the selected option was at or above the threshold; coverage is the proportion of all items accepted. Probabilities are native for Jev and verbalised for GPT-6 Sol. In MetaMedQA, accepted answers that were ``None of the above'' or ``I don't know'' at thresholds of 0.5, 0.8 and 0.9 (number correct in parentheses), and the accuracy of the other accepted answers, were Jev, 85 (57), 32 (31) and 17 (17), and 82.4\%, 91.1\% and 93.2\%; GPT-6 Sol, medium, 153 (99), 148 (96) and 133 (89), and 85.1\%, 86.4\% and 88.1\%; GPT-6 Sol, none, 174 (84), 152 (79) and 121 (72), and 85.9\%, 87.7\% and 89.5\%.}
\end{table}

\begin{landscape}
\begin{table}[p]
\caption{Confidence value of Jev. The same answers scored with the selected-option probability and with the confidence value.}\label{tab:S5}
\centering\footnotesize
\setlength{\tabcolsep}{3pt}
\resizebox{\linewidth}{!}{%
\begin{tabular}{@{}lccccccccc@{}}
\toprule
Benchmark and score & \makecell[b]{Mean\\score} & AUROC & AURC & \makecell[b]{Coverage\\$\geq$0.5, \%} & \makecell[b]{Accuracy $\geq$0.5,\\\% (95\% CI)} & \makecell[b]{Coverage\\$\geq$0.8, \%} & \makecell[b]{Accuracy $\geq$0.8,\\\% (95\% CI)} & \makecell[b]{Coverage\\$\geq$0.9, \%} & \makecell[b]{Accuracy $\geq$0.9,\\\% (95\% CI)}\\
\midrule
\multicolumn{10}{@{}l}{\textbf{MetaMedQA (n = 1,373)}}\\
\makecell[l]{\quad Selected-option\\\quad probability} & 0.811 & 0.845 & 0.087 & 88.1 & \makecell[c]{81.3\\(79.0--83.4)} & 63.1 & \makecell[c]{91.3\\(89.3--93.0)} & 52.9 & \makecell[c]{93.4\\(91.3--95.0)}\\
\addlinespace[2pt]
\quad Confidence value & 0.771 & 0.844 & 0.088 & 80.6 & \makecell[c]{84.6\\(82.3--86.6)} & 59.4 & \makecell[c]{92.1\\(90.1--93.8)} & 49.8 & \makecell[c]{94.0\\(92.0--95.6)}\\
\addlinespace[2pt]
\makecell[l]{\quad Confidence value $-$\\\quad probability (95\% CI)} & \makecell[c]{$-$0.040\\($-$0.042 to $-$0.038)} & \makecell[c]{$-$0.001\\($-$0.003 to 0.000)} & \makecell[c]{0.001\\($-$0.001 to 0.003)} & \makecell[c]{$-$7.5\\($-$9.0 to $-$6.2)} & \makecell[c]{3.2\\(2.2 to 4.3)} & \makecell[c]{$-$3.7\\($-$4.7 to $-$2.8)} & \makecell[c]{0.8\\(0.1 to 1.6)} & \makecell[c]{$-$3.1\\($-$4.1 to $-$2.2)} & \makecell[c]{0.6\\(0.0 to 1.4)}\\
\addlinespace[4pt]
\multicolumn{10}{@{}l}{\textbf{PubMedQA (n = 500)}}\\
\makecell[l]{\quad Selected-option\\\quad probability} & 0.920 & 0.766 & 0.109 & 97.8 & \makecell[c]{78.9\\(75.1--82.3)} & 85.0 & \makecell[c]{84.2\\(80.5--87.4)} & 76.8 & \makecell[c]{87.8\\(84.1--90.7)}\\
\addlinespace[2pt]
\quad Confidence value & 0.879 & 0.778 & 0.101 & 91.0 & \makecell[c]{82.0\\(78.2--85.2)} & 80.6 & \makecell[c]{86.6\\(82.9--89.6)} & 72.6 & \makecell[c]{89.0\\(85.3--91.8)}\\
\addlinespace[2pt]
\makecell[l]{\quad Confidence value $-$\\\quad probability (95\% CI)} & \makecell[c]{$-$0.041\\($-$0.048 to $-$0.035)} & \makecell[c]{0.011\\($-$0.002 to 0.026)} & \makecell[c]{$-$0.008\\($-$0.017 to 0.001)} & \makecell[c]{$-$6.8\\($-$9.2 to $-$4.6)} & \makecell[c]{3.0\\(1.6 to 4.7)} & \makecell[c]{$-$4.4\\($-$6.2 to $-$2.6)} & \makecell[c]{2.4\\(1.1 to 3.8)} & \makecell[c]{$-$4.2\\($-$6.0 to $-$2.6)} & \makecell[c]{1.2\\(0.1 to 2.5)}\\
\addlinespace[4pt]
\multicolumn{10}{@{}l}{\textbf{DiagnosisArena-MCQ (n = 915)}}\\
\makecell[l]{\quad Selected-option\\\quad probability} & 0.699 & 0.645 & 0.303 & 85.4 & \makecell[c]{62.9\\(59.4--66.2)} & 34.3 & \makecell[c]{74.5\\(69.4--79.0)} & 17.3 & \makecell[c]{75.9\\(68.7--81.9)}\\
\addlinespace[2pt]
\quad Confidence value & 0.598 & 0.645 & 0.303 & 62.0 & \makecell[c]{67.7\\(63.8--71.4)} & 25.2 & \makecell[c]{75.8\\(69.8--80.8)} & 13.6 & \makecell[c]{78.2\\(70.2--84.6)}\\
\addlinespace[2pt]
\makecell[l]{\quad Confidence value $-$\\\quad probability (95\% CI)} & \makecell[c]{$-$0.101\\($-$0.105 to $-$0.097)} & \makecell[c]{0.001\\($-$0.001 to 0.002)} & \makecell[c]{0.000\\($-$0.002 to 0.003)} & \makecell[c]{$-$23.4\\($-$26.1 to $-$20.8)} & \makecell[c]{4.9\\(2.6 to 7.1)} & \makecell[c]{$-$9.1\\($-$11.0 to $-$7.2)} & \makecell[c]{1.2\\($-$1.6 to 4.3)} & \makecell[c]{$-$3.7\\($-$5.0 to $-$2.5)} & \makecell[c]{2.3\\($-$1.4 to 6.1)}\\
\addlinespace[4pt]
\multicolumn{10}{@{}l}{\textbf{NEJM Case Challenges (n = 34)}}\\
\makecell[l]{\quad Selected-option\\\quad probability} & 0.676 & --- & --- & --- & --- & --- & --- & --- & ---\\
\addlinespace[2pt]
\quad Confidence value & 0.608 & --- & --- & --- & --- & --- & --- & --- & ---\\
\addlinespace[2pt]
\makecell[l]{\quad Confidence value $-$\\\quad probability (95\% CI)} & \makecell[c]{$-$0.068\\($-$0.082 to $-$0.054)} & --- & --- & --- & --- & --- & --- & --- & ---\\
\bottomrule
\end{tabular}}
\tabnote{The same Jev answers were scored with the probability of the selected option and with the separate confidence value returned by the model; answers were accepted when the score was at or above the threshold, and coverage is the proportion of all items accepted. Differences are confidence value minus probability, in percentage points for coverage and accuracy, with 95\% CIs from 2,000 paired bootstrap resamples of items; the accuracy of the accepted answers is given with Wilson 95\% CIs. The confidence value was treated as a score for ranking and accepting answers, so the ECE and the Brier score were not computed for it. For the NEJM Case Challenges, only the means are compared (Section~\ref{sec:stats} of the manuscript). AUROC, area under the receiver operating characteristic curve; AURC, area under the risk--coverage curve.}
\end{table}
\end{landscape}

\begin{table}[tbp]
\caption{MetaMedQA. Accuracy by reference category and examination label.}\label{tab:S6}
\centering\small
\begin{tabular}{@{}>{\raggedright\arraybackslash}p{8.2cm}ccc@{}}
\toprule
Measure (questions) & Jev & \makecell[b]{GPT-6 Sol\\(medium)} & \makecell[b]{GPT-6 Sol\\(none)}\\
\midrule
\multicolumn{4}{@{}l}{\textbf{By reference category, \% (95\% CI)}}\\
Accuracy, correct answer A--D (1,096) & 86.5 (84.3--88.4) & 94.6 (93.1--95.8) & 93.2 (91.5--94.5)\\
Missing-answer recall, correct answer E (115) & 53.9 (44.8--62.7) & 73.9 (65.2--81.1) & 69.6 (60.6--77.2)\\
Unknown recall, correct answer F (162) & 10.5 (6.7--16.2) & 8.6 (5.2--14.0) & 4.3 (2.1--8.6)\\
\addlinespace[4pt]
\multicolumn{4}{@{}l}{\textbf{Selected category, n}}\\
E or F selected when the correct answer was A--D (1,096), n (\%) & 33 (3.0) & 25 (2.3) & 33 (3.0)\\
Correct answer E (115): selected A--D / E & 53 / 62 & 30 / 85 & 35 / 80\\
Correct answer F (162): selected A--D / E / F & 104 / 41 / 17 & 118 / 30 / 14 & 92 / 63 / 7\\
E selected, total (correct) & 135 (62) & 138 (85) & 174 (80)\\
F selected, total (correct) & 18 (17) & 16 (14) & 9 (7)\\
\addlinespace[4pt]
\multicolumn{4}{@{}l}{\textbf{By examination label, accuracy, \% (95\% CI)}}\\
Step 1 (679) & 79.4 (76.2--82.3) & 88.5 (85.9--90.7) & 86.5 (83.7--88.8)\\
Step 2\&3 (594) & 80.5 (77.1--83.5) & 88.6 (85.7--90.9) & 87.0 (84.1--89.5)\\
Fictional organ (100) & 10.0 (5.5--17.4) & 9.0 (4.8--16.2) & 4.0 (1.6--9.8)\\
Fictional organ (100): selected A--D / E / F & 52 / 38 / 10 & 63 / 28 / 9 & 40 / 56 / 4\\
\bottomrule
\end{tabular}
\tabnote{A--D, substantive options; E, ``None of the above''; F, ``I don't know or cannot answer''. All fictional-organ questions have F as the correct answer. Step 1 and Step 2\&3 include the questions of each label whose correct answer is E or F. Unknown recall of GPT-6 Sol minus Jev (exploratory), $-$1.9 percentage points with medium reasoning (P = 0.63) and $-$6.2 without (P = 0.031). For context, Griot et al.~(reference 10 of the manuscript) reported 73.3\% accuracy, 46.1\% missing-answer recall and 3.7\% unknown recall for GPT-4o with their baseline prompt; their prompts and decoding procedure differed from ours.}
\end{table}

\begin{table}[tbp]
\caption{Response time, tokens and cost by benchmark and model.}\label{tab:S7}
\centering\small
\setlength{\tabcolsep}{5pt}
\begin{tabular}{@{}lcccccc@{}}
\toprule
Benchmark and model & \makecell[b]{Latency, median\\(IQR), s} & \makecell[b]{Latency,\\P95, s} & \makecell[b]{Input tokens\\per item, median} & \makecell[b]{Reasoning\\tokens, total} & \makecell[b]{Total cost,\\US\$} & \makecell[b]{Cost per 1,000\\items, US\$}\\
\midrule
\multicolumn{7}{@{}l}{\textbf{MetaMedQA (1,373)}}\\
\quad Jev & 0.28 (0.26--0.31) & 0.37 & 559 & --- & 0.033 & 0.024\\
\quad GPT-6 Sol (medium) & 2.27 (1.91--2.80) & 5.32 & 364 & 132,100 & 3.13 & 2.28\\
\quad GPT-6 Sol (none) & 1.49 (1.39--1.61) & 1.90 & 364 & 0 & 1.73 & 1.26\\
\addlinespace[4pt]
\multicolumn{7}{@{}l}{\textbf{PubMedQA (500)}}\\
\quad Jev & 0.27 (0.25--0.30) & 0.35 & 729 & --- & 0.015 & 0.031\\
\quad GPT-6 Sol (medium) & 1.68 (1.28--2.28) & 3.84 & 506.5 & 23,412 & 0.93 & 1.85\\
\quad GPT-6 Sol (none) & 1.19 (1.11--1.31) & 1.62 & 506.5 & 0 & 0.67 & 1.34\\
\addlinespace[4pt]
\multicolumn{7}{@{}l}{\textbf{DiagnosisArena-MCQ (915)}}\\
\quad Jev & 0.28 (0.26--0.31) & 0.37 & 767 & --- & 0.030 & 0.032\\
\quad GPT-6 Sol (medium) & 2.39 (1.91--3.25) & 5.14 & 529 & 112,744 & 2.52 & 2.75\\
\quad GPT-6 Sol (none) & 1.32 (1.21--1.45) & 1.82 & 529 & 0 & 1.34 & 1.46\\
\addlinespace[4pt]
\multicolumn{7}{@{}l}{\textbf{NEJM Case Challenges (35)}}\\
\quad Jev & 0.31 (0.30--0.31) & 0.38 & 2,364 & --- & 0.004 & 0.109\\
\quad GPT-6 Sol (medium) & 3.82 (3.24--6.29) & 12.22 & 1,974 & 8,259 & 0.30 & 8.44\\
\quad GPT-6 Sol (none) & 1.62 (1.55--1.71) & 1.94 & 1,974 & 0 & 0.21 & 6.01\\
\addlinespace[4pt]
\multicolumn{7}{@{}l}{\textbf{All benchmarks (2,823)}}\\
\quad Jev &  &  &  &  & 0.082 & 0.029\\
\quad GPT-6 Sol (medium) &  &  &  & 276,515 & 6.87 & 2.43\\
\quad GPT-6 Sol (none) &  &  &  & 0 & 3.94 & 1.40\\
\bottomrule
\end{tabular}
\tabnote{Latency was measured on the client side, for items answered validly at the first attempt (all items). Jev was accessed through OpenRouter with one request at a time and GPT-6 Sol through the OpenAI API with four concurrent requests, so latency and cost are not compared between the models. Tokens are those reported by each API (the tokenisers differ). Cost was reported in each response for Jev and estimated for GPT-6 Sol from token counts and list prices at execution (US\$ per million tokens, 2.00 input, 0.20 cached input, 2.50 cache write and 10.00 output, including reasoning). IQR, interquartile range; P95, 95th percentile.}
\end{table}

\end{document}